\documentclass[conference]{IEEEtran}
\IEEEoverridecommandlockouts
\usepackage{cite}
\usepackage{amsmath,amssymb,amsfonts}
\usepackage{algorithmic}
\usepackage{graphicx}
\usepackage{textcomp}
\usepackage{xcolor}
\def\BibTeX{{\rm B\kern-.05em{\sc i\kern-.025em b}\kern-.08em
    T\kern-.1667em\lower.7ex\hbox{E}\kern-.125emX}}
\begin{document}

\title{
Proposing a Game Theory Approach to Explore Group Dynamics with Social Robots\\

}

\author{\IEEEauthorblockN{1\textsuperscript{st} Giulia Pusceddu}
\IEEEauthorblockA{\textit{Italian Institute of Technology} \\
Genoa, Italy \\
0000-0002-5426-5448}
}

\maketitle

\begin{abstract}
Integrating social robots in our group-based society, beyond the technical challenges, requires considering the social group dynamics. 
Following the results from preliminary exploratory studies on the influence of social robots on group decisions, the proposed research investigates whether social robots can foster cooperation among group members.
To achieve this, I propose a game theory approach, employing the Public Good Game to recreate a simplified and controlled social situation where the robot's influence can be evaluated.  
Clarifying the role of robots in promoting collaboration among humans might have a significant impact in educational environments, enhancing student learning, as well as in workplace settings, where they could facilitate problem-solving and lead to shared solutions.


\end{abstract}

\begin{IEEEkeywords}
group-robot interaction, multi-party interaction, group dynamics, game theory 
\end{IEEEkeywords}

\section{Introduction \& Motivation}
Since the earliest times, humans have structured their society around groups, from close-knit contexts such as families and friendships to goal-oriented organizations like work and sports teams, with the result that most human activities today happen within groups. 
If the Human-Robot-Interaction (HRI) community aspires to develop social robots that, in the future, will support humans in their daily lives, it is essential to equip them with the ability to deal with groups of people.
Achieving this goal is surely challenging, not only due to the technical obstacles
but also because of the social science aspects that need to be taken into consideration. 
Indeed, for a social robot, interacting with a group should not merely mean replicating one-to-one interactions with every member.
The concept of groups entails a complex set of group dynamics - i.e., interpersonal processes that occur in groups \cite{forsyth2014group} - that, ideally, robots should take into account, as humans do. 

I believe that social robotic agents can become valuable support to guide teams toward good performance, cooperation, and satisfaction of their members.
For this reason, in my research on group dynamics, I have begun to investigate whether social robots may impact group decisions and collaboration. 

So far, it is unclear whether the influence of robots in decision-making scenarios within groups is similar to that exerted by humans. 
Studies using variations of the Asch experiment have found contradictory results \cite{vollmer2018children, salomons2018humans, qin2022adults, brandstetter2014peer}.
Other research suggests that robot advice is only considered when evaluated low-impact \cite{sembroski2017he}, and that human members tend to prefer following human members' advice over robots' \cite{zhang2021you}. 
Additionally, research that examines the robot’s role in a group suggests that even when robots are introduced as leaders, humans might not perceive them as such \cite{alves2016role}. 

Previous studies have shown that in conversations robots can use verbal or nonverbal behaviors, such as gaze direction, to encourage individuals who typically engage less, helping to keep group interactions balanced \cite{neto2023robot, grassi2023robot, gillet2021robot}.
While preliminary studies suggest that social robots may promote collaboration in groups of children \cite{strohkorb2016improving, gillet2020social}, it remains unclear whether robots can actively foster cooperation.

With my research, I aim to explore these aspects of group dynamics further by addressing two research questions.
First, I have been seeking to determine whether and to what extent 
a social robot can influence other members' decisions (\textbf{Q1}). 
Since the beginning of my PhD, I have explored and tested different experimental methods to understand under what conditions a social robot could leverage in its team's choices (Section \ref{preliminary}).
Second, after finding encouraging results, I decided to take it a step further and focus on determining whether social robots can promote group cooperation (\textbf{Q2}).
To advance in this direction, more controlled experiments and rigorous evaluation methods are essential. 
Game theory presents a particularly promising approach for this purpose, as it considers both individual benefits and the overall group context and has already been used to explore complex social behaviors. 
I propose to use this approach to design measurable tasks that replicate simplified social scenarios within which I will investigate my second research question (Section \ref{proposed}). 

\section{Game Theory in HRI}
Game theory was initially designed to investigate firms' and consumers' behavior in the economic field \cite{von2007theory}. 
It provides a mathematical framework to examine how people make decisions when their actions have an effect on others, and, at the same time, their fates depend on the actions of other agents. 
Its nature makes it suitable for investigating social behavior, with particular mention of scenarios in which: 
(i) joint strategies cannot be pre-planned (non-cooperative), and 
(ii) all players can achieve a positive score (non-zero-sum).
Non-zero-sum non-cooperative games,
such as Prisoner's Dilemma and Public Good Game,  
have been extensively used in social sciences to investigate cooperation and competition among humans \cite{sigmund2010social, dong2016dynamics, chaudhuri2002cooperation}.  

In the last years, even HRI researchers have started employing game theory to evaluate cooperation and reciprocity between humans and artificial agents, mostly in dyadic interactions.
Experiments based on games like the Ultimatum Game have unveiled that humans tend to be equivalently reciprocal with humans and robots \cite{sandoval2016reciprocity, sandoval2021robot}, and this tendency might even surpass the influence of the reward value of their decisions \cite{hsieh2020human}.
These results are promising because, according to the definition of reciprocity \cite{falk2006theory}, they indicate that humans attribute to robots the faculty of responding cooperatively to kind actions and harshly to hostile ones.
Although the same studies have shown that humans tend to cooperate more with other humans than with robotic agents, research using a modified version of the Public Good Game has demonstrated that robots displaying cooperative behavior are preferred over those that defect, regardless of the overall game score \cite{correia2019exploring}.

I believe that game theory, which has mostly been applied in one-on-one scenarios in HRI, still holds relevant potential for studying group dynamics. 
Given the extensive use of game theory made by the social sciences, this approach offers a solid framework for comparing interactions in human-only versus human-robot mixed groups. 
Furthermore, decision games such as the Public Good Game, recreate simplified social scenarios where quantitative measures of players' benefits and strategies can be analyzed.
For these reasons, I aim to apply a decision game to understand whether social robots, through their behavior, can foster collaboration within a team (Q2).

\section{Preliminary Work}\label{preliminary}
To evaluate whether a robot could influence group decisions (Q1) I started testing groups of children during a team collaborative strategy game \cite{hidden2025exploring}. 
The robot, presented as the leader of the interaction, advised on game moves, and the participants chose whether to follow or ignore its tips.
Participants rarely decided to follow the robot's advice, especially after some of them turned out to be wrong. 
A significant association between the strategies of the players who consistently initiated moves (i.e., ``initiators'') and their teammates was found. 
Players seemed to conform to the initiator's strategy, whether or not it was in line with the robot's advice. 
Interestingly, these results were found using both Nao and iCub, thus suggesting that this dynamic does not depend on the type of robotic platform. 

Similar effects were also found in another study, in which children had to mimic actions before and after watching the robot Nao perform them \cite{hidden2023the}. Participants tended not to imitate the robot if it performed moves atypically. 
However, again it was found that the more proactive players in the group tended to influence the actions of others: players were significantly more likely to mimic the robot when the initiator conformed to Nao.
These findings not only reinforce the notion that taking the initiative is a leadership trait \cite{yoo2004emergent}, but also suggest that in these contexts, the robot's influence on a group's choices might be mediated by the behavior of an initiator.

I further explored decision-making (Q1) with adults, testing whether two Furhats could influence the choices of a human in a task in which they had to label ambiguous AI-generated images.
Participants rarely changed their responses to align with the robots, even when both disagreed with their choice; though confirmation took longer in these cases, indicating possible indecision and thus that robots may have the potential to prompt reconsideration in human judgment.

\section{Proposed Work}\label{proposed}
After my preliminary studies,
I decided to continue exploring the influence of social robots on groups making two adjustments: (1) removing the mediation effect of the initiators, and (2) concentrating on a more specific aspect of decision-making: collaboration among group members (Q2). 

To pursue my goal, I chose to employ a game theory task called Public Good Game, in which group members play simultaneously. 
In this game, during each round, individuals decide how much of their private resources to contribute to a common pool. 
Total contributions are then multiplied by a bonus and evenly distributed, among all participants, 
regardless of their individual input, creating a trade-off between self-interest and collective benefit.

Research testing the Public Good Game with only human players has found that the presence of cooperating members promotes higher average group contributions \cite{wu2014role}. This study will verify if this effect can also be elicited by a robotic agent. 

To test these dynamics, the proposed study will involve a 50-round Public Good Game, with two human participants and a social robot across two conditions.
In the first, the robot will exhibit cooperative behavior by contributing an amount equal to or greater than the average contributions of the previous round to the common pool. 
In the second condition, the robot's cooperative behavior will be accompanied by social cues, such as verbal feedback and facial expressions. This addition aims to determine whether merely exhibiting cooperative behavior is sufficient to foster collaboration, or if the integration of social cues by a social robot may cause or amplify cooperation.
Furthermore, the repeated rounds of the game will reveal how players’ strategies develop and adapt over time. 

\section{Contribution}
Gaining more insight into how the presence of a robot can enhance group collaboration could have a significant impact in various contexts. 
For instance, in the field of educational robotics, it could improve the performance of student groups \cite{strohkorb2016improving}. 
In the workplace, robots could support problem-solving within teams or assist in negotiations during conflict situations to reach a shared solution \cite{jung2015using}.

To investigate the complex dynamics of collaboration, I propose a methodology based on game theory that allows for meaningful comparisons with research conducted on human-only groups. 
The framework allows the exploration of various robot strategies, besides the cooperating approach proposed. 
Additionally, it permits flexibility in testing different quantities of human or robotic players.
Building on this approach, future work could explore additional group dynamics, including leader-follower patterns and the development of group trust, possibly taking a step closer to integrating robots into our group-based society.

\bibliographystyle{ieeetr}
\bibliography{bibliography}

\end{document}